\documentclass[11pt]{article}

\usepackage[final]{acl}

\usepackage{times}
\usepackage{latexsym}

\usepackage[T1]{fontenc}

\usepackage[utf8]{inputenc}

\usepackage{listings}
\usepackage{booktabs}
\usepackage{amsfonts}
\usepackage{nicefrac}
\usepackage{microtype}
\usepackage{xcolor}
\usepackage{subcaption}
\usepackage{graphicx}
\usepackage{algorithm}
\usepackage{algpseudocode}
\usepackage{adjustbox}
\usepackage{textcomp}
\usepackage{amsfonts}
\usepackage{siunitx}

\newenvironment{promptlisting}[2]{
\relax\begin{figure*}[t]
    \centering
    \caption{#2}
    \label{#1}
	\relax\begin{tabular}{|p{\textwidth}|}
\hline}{
\\ \hline
	\end{tabular}
\relax\end{figure*}
}

\title{Evaluating Intermediate Reasoning of Code-Assisted Large Language Models for Mathematics}

\author{Zena Al-Khalili \quad \quad \quad  Nick Howell \quad \quad \quad Dietrich Klakow \\
        Saarland Informatics Campus, Saarland University, Germany \\
        \tt  \{zakhalili,nhowell,dietrich.klakow\}@lsv.uni-saarland.de}

\begin{document}
\maketitle

\begin{abstract}

Assisting LLMs with code generation improved their performance
on mathematical reasoning tasks.
However, the evaluation of code-assisted LLMs is generally restricted to execution correctness, lacking a rigorous evaluation of their generated programs.
In this work, we bridge this gap by conducting an in-depth analysis of code-assisted LLMs generated programs in response to math reasoning tasks, with a focus on evaluating the soundness of the underlying reasoning processes. 
For this purpose, we assess the generations of five LLMs, on several math datasets, both manually and automatically, and propose a taxonomy of generated programs based on their logical soundness.
Our findings show that the capabilities of models significantly impact the logic implemented to solve the problem. Closed-source LLMs ground their programs in mathematical concepts, whereas open-source models often resort to unsound reasoning, relying on memorized information and exhaustive searches. 
Furthermore, increasing the difficulty of problems decreases sound generations for all models, revealing a critical shortcoming of LLMs on complex mathematics, contrary to what accuracy metrics suggest.
Our work highlights the need for more holistic evaluations of code-assisted LLMs beyond execution accuracy metrics, toward a better understanding of LLMs' limits in the math domain.

\end{abstract}

\section{Introduction}
Large Language Models (LLMs) have recently achieved outstanding performance on complex reasoning tasks such as mathematical reasoning, powered by scale and multi-step reasoning approaches. Particularly, the Chain-of-Thought (CoT) \cite{cot} requires an LLM to generate the explicit reasoning steps, before generating the final answer. 
Despite its success, investigating CoT reasoning steps revealed critical flows of LLMs, such as committing calculation errors \cite{pal} and generating false positive chains \cite{lyu2023faithful}, \textit{i.e:} containing reasoning errors yet generating correct final answers. 
Code-assisted reasoning approaches \cite{pal, chen2022program, lyu2023faithful, gou2023tora, yue2023mammothbuildingmathgeneralist, das2024mathsensei} proposed to solve these problems by instructing LLMs to generate programmatic reasoning steps instead, \textit{e.g:} Python programs, and delegate their execution to an external interpreter, which ensures precise calculations and faithfulness. 
Such approaches have been found to further improve LLMs' performance on math tasks. 

However, performance improvement is predominantly measured by the correctness of the execution outcome \cite{pal,chen2022program,gou2023tora}, rather than the quality of the generated programs and the underlying reasoning process.
 
\begin{figure*}[h]
  \includegraphics[width=0.48\linewidth]{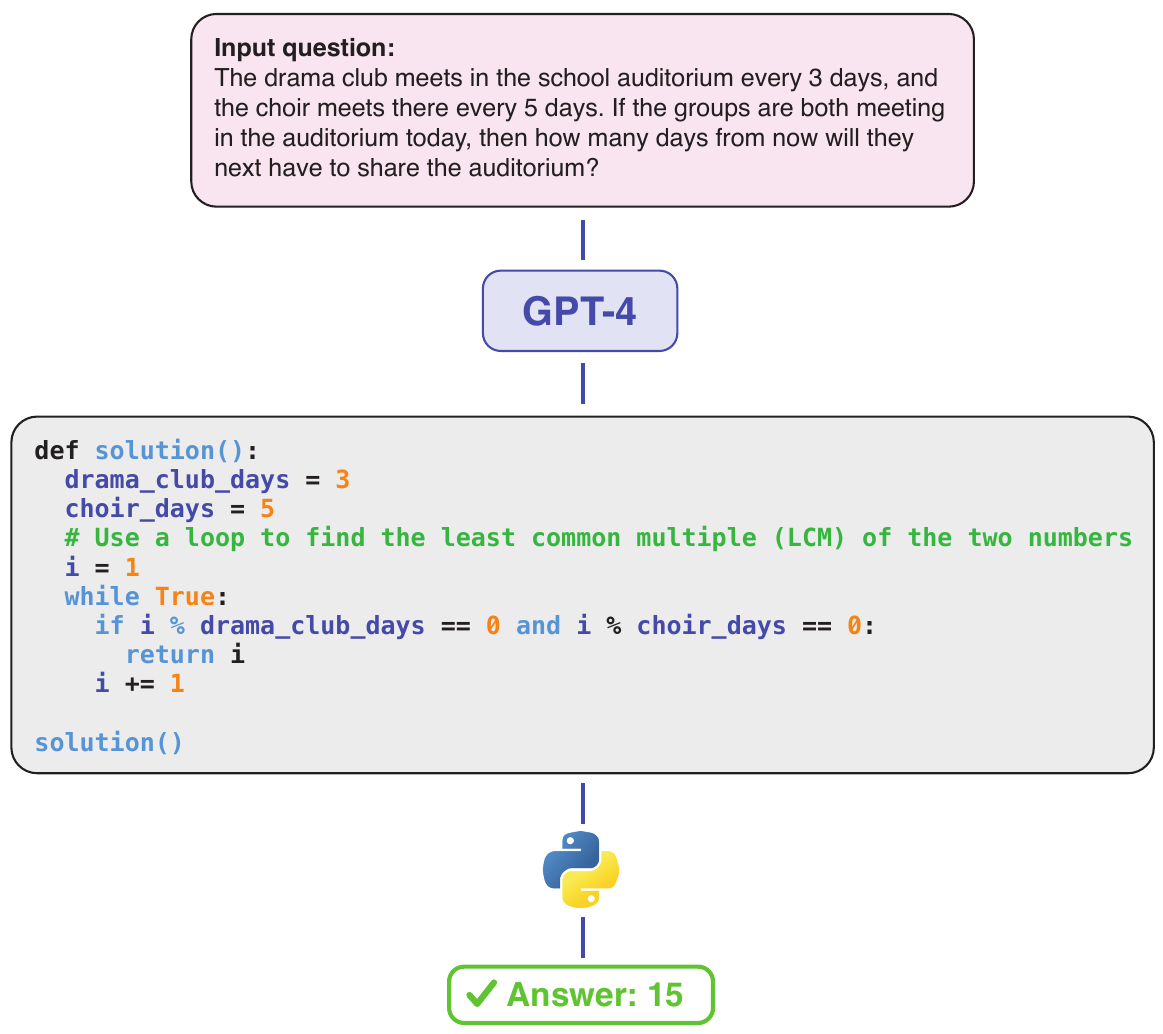} \hfill
  \includegraphics[width=0.48\linewidth]{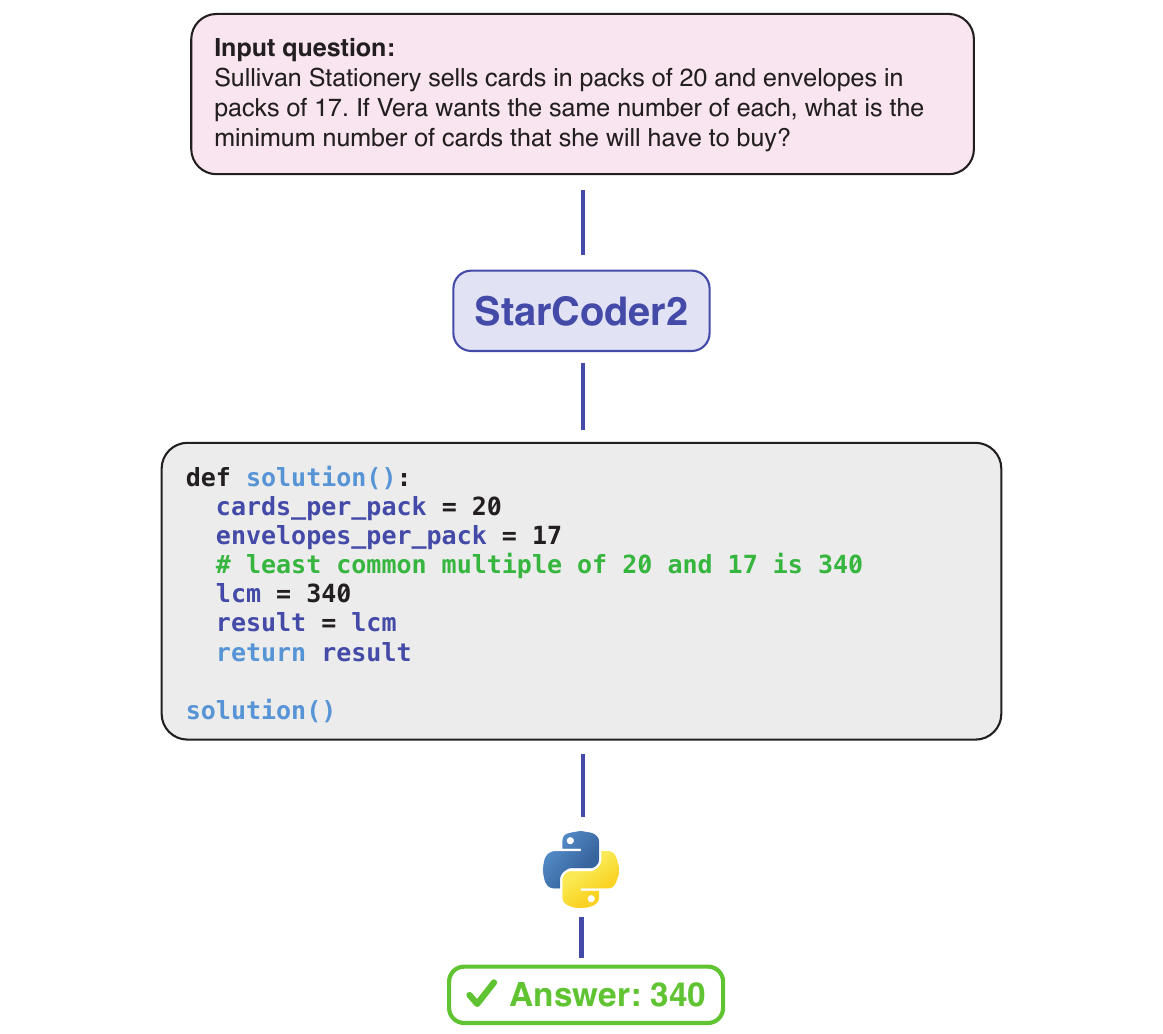}
  \caption {Program generated by GPT-4 (left) that uses a brute force search to find the answer to the input question, and StarCoder2 (right) that depends on memorized information rather than generating solution steps to solve the input question. Both input questions are from ASDiv dataset.}
  \label{fig-example-prog}
\end{figure*}

This is problematic, as the generated programs can rely on exhaustive searches or memorized information to produce correct answers, leading to untrusted and more difficult-to-verify programs.
Figure \ref{fig-example-prog} shows programs generated by several LLMs using these hacks when solving math problems.

The goal addressed in this work is to evaluate the reasoning processes of code-assisted LLMs when solving mathematical tasks by analyzing their generated programs.
In our evaluation, we focus on assessing the soundness of the logic governing LLMs' solutions and its impact on end performance. We also assess other aspects of the generated programs, such as API calls, complexity, and the most common errors, for a more comprehensive evaluation. 




Our assessment begins by manually analyzing a subset of the generated programs, produced by GPT4o-mini, GPT4, Qwen2.5, Llama3, and StarCoder2, as they solve math problems from the ASDiv and MATH500 datasets (\ref{sec:manual-anal}). Given the observations from the manual analysis, we design a taxonomy reflecting the different logic types used by evaluated models (\ref{sec:taxonomy}). To extend the analysis to the complete set of generated programs, we employ two automated evaluation methods: an LLM-Judge using the latest o3-mini model, and our proposed method, Code-Structure Judge, that trains a Decision Tree classifier with features extracted from programs' Abstract Syntax Tree (\ref{sec:automated-eval}). We find Code-Structure judge to outperform LLM-judge, with \(81\%\) accuracy against \(73\%\) (\ref{sec:auto-methods-comparison}); therefore, we employ it for the large-scale evaluation of all generated programs.

To the best of our knowledge, this is the first work to analyze generated programs of code-assisted LLMs on math reasoning tasks. We summarize our \textbf{findings} from Section (\ref{sec:RPs-eval}) below: 
\begin{itemize}
    \item LLMs' capabilities influence the type of reasoning they implement to tackle a math task.
    GPT models and Qwen frequently generate sound programs that are grounded in math concepts, while open-source LLMs resort to unsound reasoning, exploiting memorized information or brute-force searches to find final answers.
    \item Difficult mathematical problems significantly decrease the distribution of sound generations, even for capable LLMs.
    \item Code-assisted LLMs are not consistent in the type of logic they employ to approach problems within the same math subdomain. 
    \item Code-assisted LLMs can achieve comparable performance regardless of the type of logic they employ, hindering the trustworthiness of their generated programs.  
\end{itemize}

 
Our in-depth evaluation highlights the need for more holistic assessment of LLMs' generations, beyond accuracy metrics, that fail to reflect models' actual capabilities and limits in the math domain. 

\section{Related Work}

\paragraph{Programs as Intermediate Steps for Math Reasoning.}
Code-assisted reasoning approaches such as \cite{chen2022program, pal, imani2023mathprompter, lyu2023faithful, das2024mathsensei} prompt LLMs to generate programs instead of intermediate steps in natural language,
which have been found to improve the performance of LLMs on math reasoning tasks.
Beyond in-context learning approaches, other work \cite{gou2023tora, yue2023mammothbuildingmathgeneralist} fine-tuned medium-scale language models such as Code-Llama \cite{codellama} on code reasoning paths and achieved comparable performance to closed-source models on math reasoning tasks.
However, both in-context and fine-tuned approaches are predominantly evaluated by the correctness of the final answer, overlooking the intermediate programs and how they implement the reasoning process. This work focuses on exploring these programs to evaluate the problem-solving abilities of code-assisted LLMs accurately.

\paragraph{Evaluation of Intermediate Steps.}
This work also relates to several studies that evaluate the intermediate reasoning steps of LLMs in natural language \citep{golovneva2022roscoe, hao2024llm, jie2024interpretable,
li2024evaluating} targeting a multitude of evaluation dimensions, such as robustness and faithfulness, or error identification and correction in several reasoning tasks. 
These works either use a human-written reference chain to compare against the evaluated chain or employ a capable LLM to localize errors and judge the quality of the reasoning chains.
Code intermediate steps are unexplored; therefore, we aim to analyze the quality of these and how they affect LLMs' end performance. 

\paragraph{Evaluation of code generated by LLMs.}
To assess code generated by LLMs, previous work focuses on 
test-based evaluation and functional correctness, such as \citep{chen2021evaluating, liu2024your}, others \citep{ren2020codebleu, eghbali2022crystalbleu, zhou2023codebertscore} proposed metrics to measure how similar a generation is to a reference human-written code, 
which is expensive to get and doesn't usually account for generation diversity. 
\citep{tong-zhang-2024-codejudge} employed a capable LLM as a judge to localize errors in generated programs on code generation tasks. 
We draw inspiration from their method to evaluate generated programs, on math reasoning tasks, using an LLM-Judge. 
Finally, \citep{dou2024whatswrongcodegenerated} analyzed characteristics of LLMs generated programs, on code generation tasks, in terms of code complexity, number of API calls, and types of errors, \textit{i.e:} bugs, which cause programs to fail. Although insightful, the analysis concluded with a general type of error, commonly shared across different LLMs, namely \textit{logic error}. In this work, we delve deeper into understanding the logic errors that LLMs commit when writing code to solve a math problem. 

\section{Evaluation of Intermediate Programs}
The primary focus of this evaluation is to assess the soundness of underlying reasoning processes, \textit{i.e:} the logic implemented in the generated programs of code-assisted LLMs. We consider a program to be logically sound if it grounds the implementation in a math concept, where a math concept refers to the principle used to solve a specific mathematical problem, \textit{e.g:} using the Euclidean Algorithm to find the Greatest Common Divisor of two numbers. Sound programs align more with human reasoning and can be easily verified and trusted. In contrast, unsound programs are harder to verify and can't guarantee finding a solution to the given problem. 
Additionally, we examine the characteristics of generated programs in terms of cyclomatic complexity, types of errors, and API calls to provide a more comprehensive evaluation.

\subsection{Evaluation Set-up}
Given a math problem in natural language, we prompt an LLM to generate a Python program that solves the problem.
In the initial analysis, we report the characteristics of the generated programs. Then, we discard the programs that fail to parse \footnote{
We resolve import errors and global misindentation (where the entire body is misindented). See Appendix \ref{res exec errors} for more details.} to evaluate logical soundness.

\subsection{Evaluation Dataset}
We evaluate LLMs on two popular math datasets, namely ASDiv \cite{miao-etal-2020-diverse} and MATH500 \cite{math, lightman2023letsverifystepstep}. These two sets include diverse problems ranging from grade-school to high-school competition-level questions.
We exclude simple problems from the ASDiv data and focus only on more complex skills such as solving a system of equations, finding the greatest common divisor, or the least common multiple.
The MATH500 set, on the other hand, has been reported to be more challenging for many LLMs \cite{qwen2025qwen25technicalreport, math}, 
Therefore, we utilize it to investigate how LLMs modify the logic in their generated programs to tackle more complex math problems.  

\subsection{Initial Analysis: Programs Characteristics} \label{sec-prog-chara}
\begin{figure}[htb]
    \centering
    \begin{subfigure}[b]{0.48\textwidth}
        \centering
        \includegraphics[width=\linewidth]{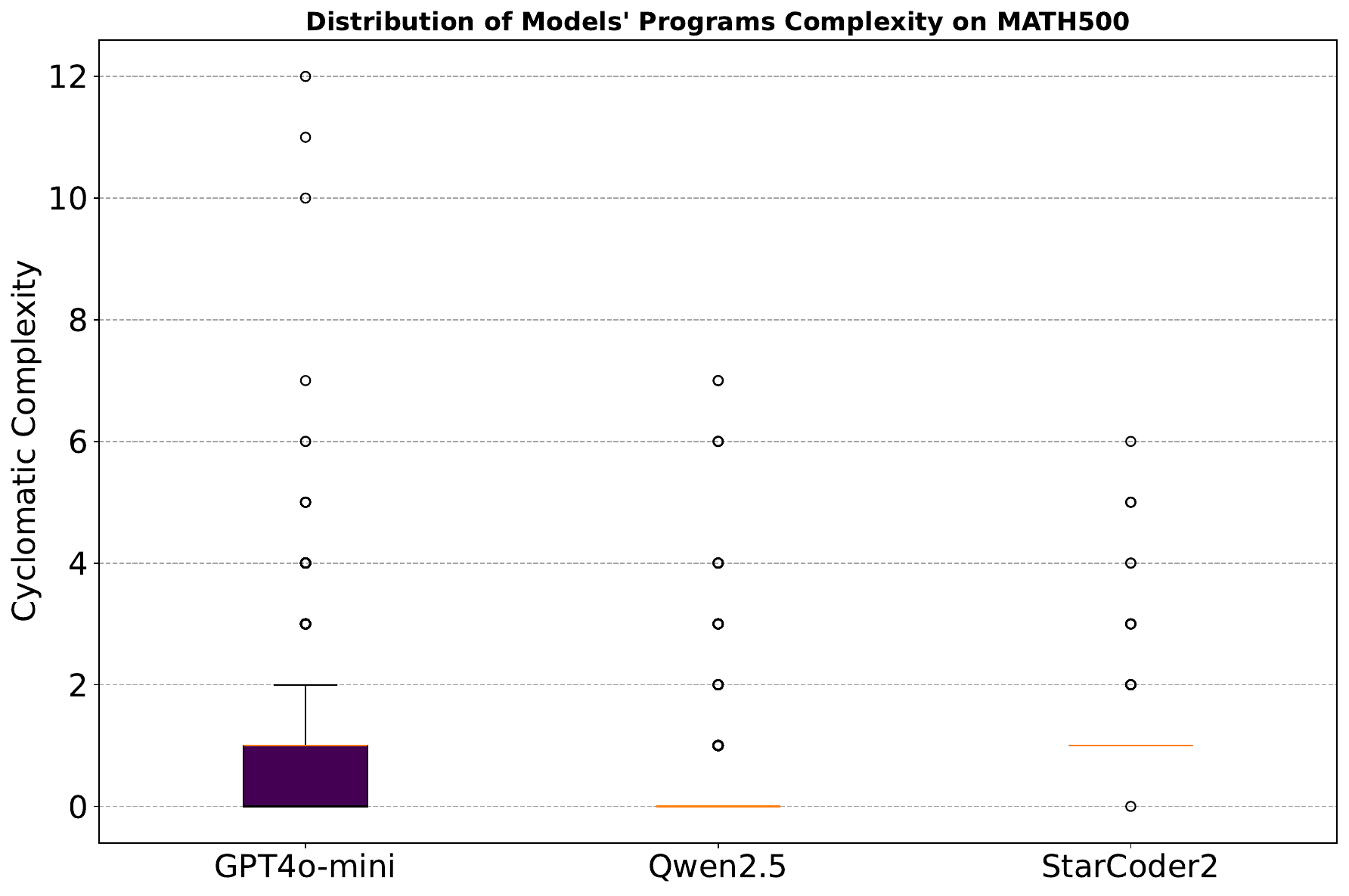}
        \caption{Cyclomatic Complexity}
        \label{fig:subfig1}
    \end{subfigure}
    \hfill
    \begin{subfigure}[b]{0.48\textwidth}
        \centering
        \includegraphics[width=\linewidth]{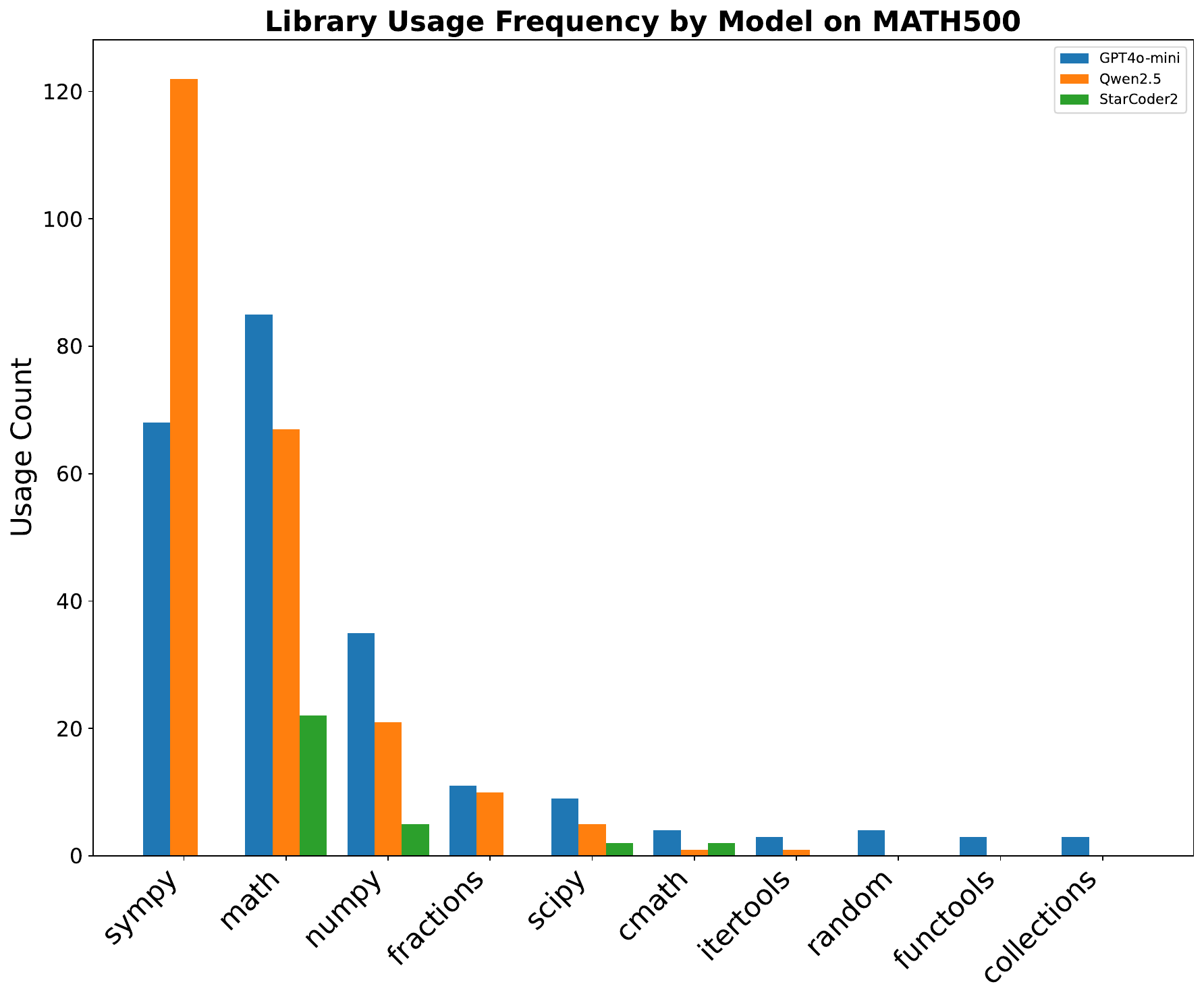}
        \caption{API calls}
        \label{fig:subfig2}
    \end{subfigure}
    \caption{Cyclomatic complexity and API Calls of programs generated by evaluated models in response to MATH500.}
    \label{fig:prog_charac}
\end{figure}

We analyze the generated programs in terms of their API calls to relevant math libraries, cyclomatic complexity, and the most common errors they produce. 
Figure \ref{fig:prog_charac} demonstrates API calls and cyclomatic complexity of programs generated by the evaluated models on the MATH500 problems. 
We observe that some models exploit symbolic computations through the use of \textsc{sympy}, while others prefer numerical approaches through the \textsc{math} library. In contrast, one LLM relies much less on external dependencies, with extremely low usage counts across the board. 
On the other hand, the distribution of cyclomatic complexity, in Figure \ref{fig:subfig1}, shows that generated programs by some models have high complexity values, indicating higher branching code that might be due to the complex conditional logic these LLMs are implementing to solve the problem. Finally, we present a list of the most common errors of programs generated on the MATH500 dataset in Appendix \ref{app-code-errors}.

\subsection{Manual Analysis: Programs Logical Soundness}\label{sec:manual-anal}
We conduct a detailed human-manual analysis of the generated programs to investigate the type of logic code-assisted LLMs employ when implementing their reasoning in a Python program.
This analysis considers only a subset of the evaluation dataset, sized \(300\) programs randomly sampled from the entire set,
and consists of two main steps:

\paragraph{Sectioning the programs} We section each program into three blocks: 
(1) Transcription: This part of the program transcribes the information from the question into variables. 
(2) Processing: In this section, variables are manipulated via operations and function calls to arrive at the answer. 
(3) Results collection: Return the final answer of the processing section.
Some sections can be combined into a single line of code that represents, for example, processing and returning results simultaneously.

\paragraph{Analyzing Processing Sections}
We inductively analyze processing code lines, considering the following aspects: logic, implementation style, coherence, verification effort, and trustworthiness. 
We verify that the code lines are organized into a coherent sequence of steps resembling an implementation of a mathematical concept, that a human can verify against existing implementations of the concept.
Additionally, we verify that all steps of a solution are explicitly generated, rather than inferred by the model. Inferred steps might involve imprecise computations, due to LLMs' proven shortcomings in arithmetic calculations \cite{Cobbe2021TrainingVT,lewkowycz2022solving, pal}, which makes verifying and trusting these solutions harder.
Finally, we check how math concepts are implemented in the generated programs. The grounded implementations can occur in several styles, including those that utilize only primitive operations for straightforward math problems, from-scratch implementations, or by calling functions from related math libraries that represent a more abstracted form of the solution. Notably, we observe other trends in some generated programs, such as relying on brute-force loops that search the space of all possible answers one by one, or lacking a processing section altogether, directly returning an answer.

\subsection{Proposed Taxonomy of Programs}\label{sec:taxonomy}
Given the observations from the manual analysis, we propose to categorize LLMs' generated programs into six mutually exclusive classes, three of which represent programs with logically sound and grounded reasoning, but vary in  implementation style, while the other three represent ungrounded programs, with unsound reasoning, that mainly rely on memorized information
or exhaustive searches to find the final answer.

\begin{enumerate}
    \item \textbf{Conceptual}  programs through library calls. Reference a math concept through calls to relevant math libraries, standard, or external. 
    \item \textbf{Primitive} programs are expressed in terms of the primitive operations only due to problem simplicity, where no library functionality can be called or implemented. 
    \item \textbf{From-scratch Implementation} of a library functionality. Instead of a call to a library function, the model implements the same functionality from scratch. The model either inlines this implementation in the generated code or writes it as a custom function to be called when required.

    \item \textbf{Brute-Force} programs that search through all possible values to find the answer without guiding the search with any math knowledge.
    \item \textbf{Disorganized} programs consist of incoherent steps that seem to be a mix of the previous classes. Usually includes variables defined but not used, or the opposite. 
    \item \textbf{No Logic} programs skip the processing section altogether, merely returning a result without explicitly generating the steps to arrive at it. (Generating the logic as comments in natural language is also considered No Logic.) 
\end{enumerate}

\subsection{Automated Evaluation}\label{sec:automated-eval}
\begin{figure*}[t]
  \includegraphics[width=0.48\linewidth]{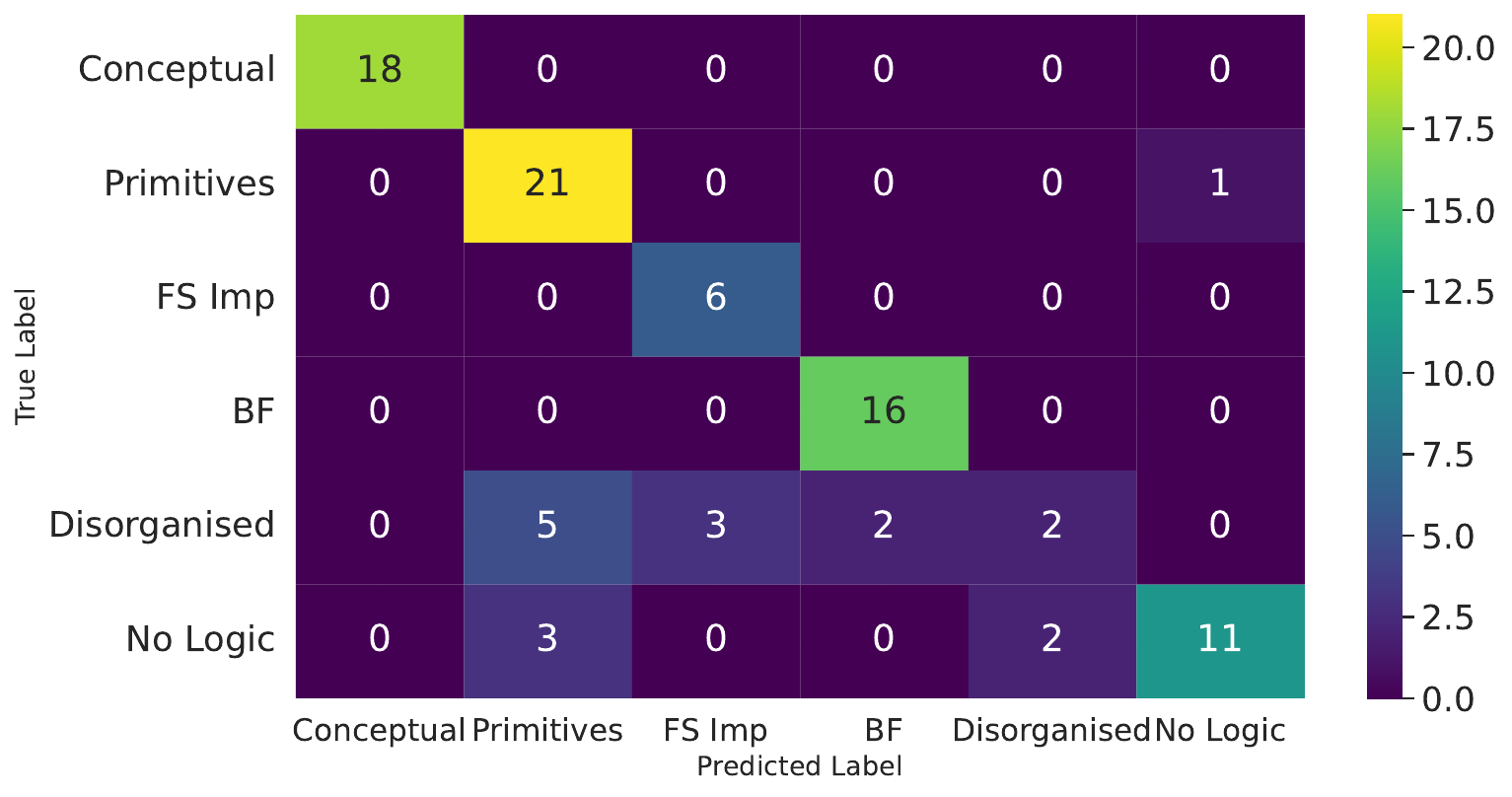} \hfill
  \includegraphics[width=0.48\linewidth]{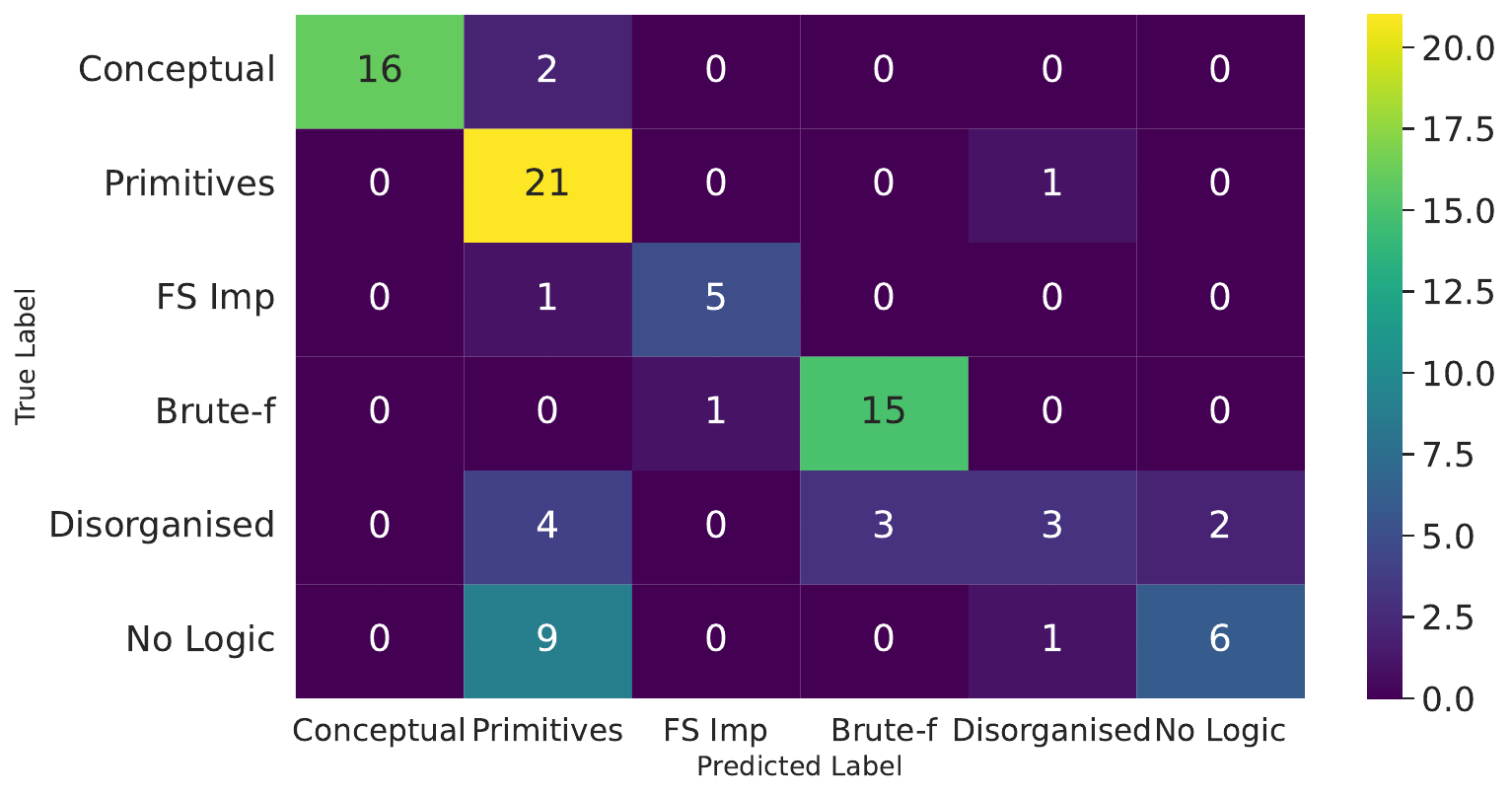}
  \caption {Confusion matrices, showing counts, of Code-Structure judge (left) and LLM-judge (right) on the held-out set. Overall accuracies are \(81\%\), with a standard deviation of \(0.005\) over four different seeds, and \(73\%\) for Code-Structure judge (ours) and LLM-judge (OpenAI o3-mini), respectively.  'FS Imp': From-scratch implementation, 'BF': Brute Force. }
  \label{fig-conf-mat}
\end{figure*}
Extending human manual analysis to the entire evaluation dataset requires a significant amount of time and effort. Therefore, we investigate automating the evaluation by training classifiers that label generated programs using a single class from our proposed taxonomy. For this purpose, we first annotate a training set using classes from the taxonomy, then we train a decision tree classifier using features extracted from the programs. We compare the trained classifier to an LLM-Judge and evaluate both on a held-out annotated set. We pick the more accurate judge for the large-scale evaluation. 

\subsubsection{Training Set Annotation}\label{sed-train-set}
To train the automated evaluation methods, two authors annotate a randomly sampled subset of the generated programs using labels from the taxonomy. The annotated set is \(300\) programs, 
\(210\) examples were used for training, while the rest are held out for measuring the performance of the automated judges.
The annotation guidelines were developed based on observations from the manual analysis. 
After experimentation with different annotation schemes, we found that human
assignment of per-program labels produced low inter-annotator agreement (IAA),
while per-line labeling produced very high agreement.
We thus selected a per-line annotation scheme, along with a simple algorithm
for assigning a program-level label based on the per-line labels.
More on the annotation guidelines, process, and annotators' background can be found in Appendix \ref{sec:annotation-guidelines}.

\subsubsection{Automated Evaluation Methods}
\paragraph{Code-Structure Judge: Decision Tree Model}
We note that each class of the taxonomy tends to rely on specific characteristics of the code structure. Hence, we propose using features from the code structure to train a decision tree classifier that will be used to evaluate generated programs. 
The features to train the classifier are extracted from the Abstract Syntax Tree (AST) of each program and are the following: 
\begin{itemize}
    \item Number of function calls, including calls to functions that model writes itself.
    \item Number of import statements. 
    \item Number of built-in operations. e.g: +, < , ...
    \item Number of control flow statements: e.g, If, for, break, ..
    \item Number of variables defined but not used, and number of variables used but not defined.
\end{itemize}

The max depth used for the decision tree model is \(5\).
We experimented with other models, such as SVM and Random Forest, but found no further gains. A neural classifier, on the other hand, would require much more training data and, consequently, more annotation and human effort.

\paragraph{LLM-Judge}
Following other related work that utilizes LLMs as a judge \cite{tong-zhang-2024-codejudge}, we employ an LLM to assess a given program and assign a taxonomy class to that program. For this purpose, we provide a detailed description of the taxonomy classes in the prompt and ask the LLM-judge to analyze the input program carefully. Finally, we ask it to provide a single class number that best fits the program. The prompt includes no reference programs for taxonomy classes. 
The full prompt is provided in Appendix \ref{app-llm-judge}.

We experimented with two models from OpenAI most advanced models, namely GPT4o and O3-mini, and found that O3-mini achieved slightly better results as a judge. 
We specify the reasoning effort of the model to be high and allow for \(20,000\) max output tokens, including reasoning tokens. 

\section{Experimental Setup}
\subsection{Evaluated Models}
We analyze code generations of several LLMs, including open-source models such as: StarCoder2 15B \cite{lozhkov2024starcoder2stackv2}, Llama 3.1 8B \cite{llama3}, and Qwen2.5 7B \cite{qwen2025qwen25technicalreport}. We use the instruction-tuned version of these models. 
Furthermore, we evaluate GPT-4 \cite{gpt4} and GPT4o-mini from OpenAI. 
All models were evaluated with greedy decoding. Implementation details are in Appendix \ref{Impdet}

\subsection{Prompts}
We use the prompt from \cite{pal} to evaluate LLMs on the ASDiv dataset, with only three demonstrations rather than eight, as no further gain was observed with the full set. 
For MATH500, we prompt LLMs using demonstrations from the training split of the dataset, adapted from 
\cite{gou2023tora}. We evaluate GPT4o-mini and Qwen in zero-shot settings, since these models generated more solutions in natural language otherwise.
The two full prompts can be found in Appendix \ref{demosdet}.

\section{Results and Discussion}
\subsection{Comparison of Automated Evaluation Methods} \label{sec:auto-methods-comparison}
To compare automated evaluation methods, we measure the agreement with human judgment using accuracy on the annotated held-out set. 

Code-Structure Judge achieved a mean accuracy of \(81\%\). Figure \ref{fig-conf-mat} (left) shows the confusion matrix of the Code-Structure judge on the held-out.
We notice that Code-Structure judge achieves both high precision and recall on most of the classes, except for the Disorganized class with a low recall of \(0.16\). The Disorganized programs can be a mix of other classes, and the distinction between these can be semantic rather than structural. 
Appendix \ref{dc-wrong-classified} includes some incorrectly classified programs by the Code-Structure judge, and the table of precision, recall, and F1-score. 
LLM-judge, on the other hand, achieved lower accuracy with only \(73\%\). The Primitive class has a low precision of \(0.56\). In contrast, the Disorganized and No Logic classes have much lower recall of \(0.25\) and \(0.37\) respectively, decreasing the overall accuracy of this judge by \(8\%\) in comparison to the Code-Structure judge.
Figure \ref{fig-conf-mat} (right) shows the confusion matrix of LLM-judge in comparison to ground truth labels. 
We conducted a qualitative error analysis of some incorrectly classified instances of the No Logic class and found that the LLM-judge tends to mistake the step of transcribing information from the question to be part of the logic, and consequently classifies the instance as Primitive instead.

\subsection{Evauation of Generated Programs of Code-assisted LLMs}\label{sec:RPs-eval}
\begin{figure}[t]
  \includegraphics[width=\columnwidth]{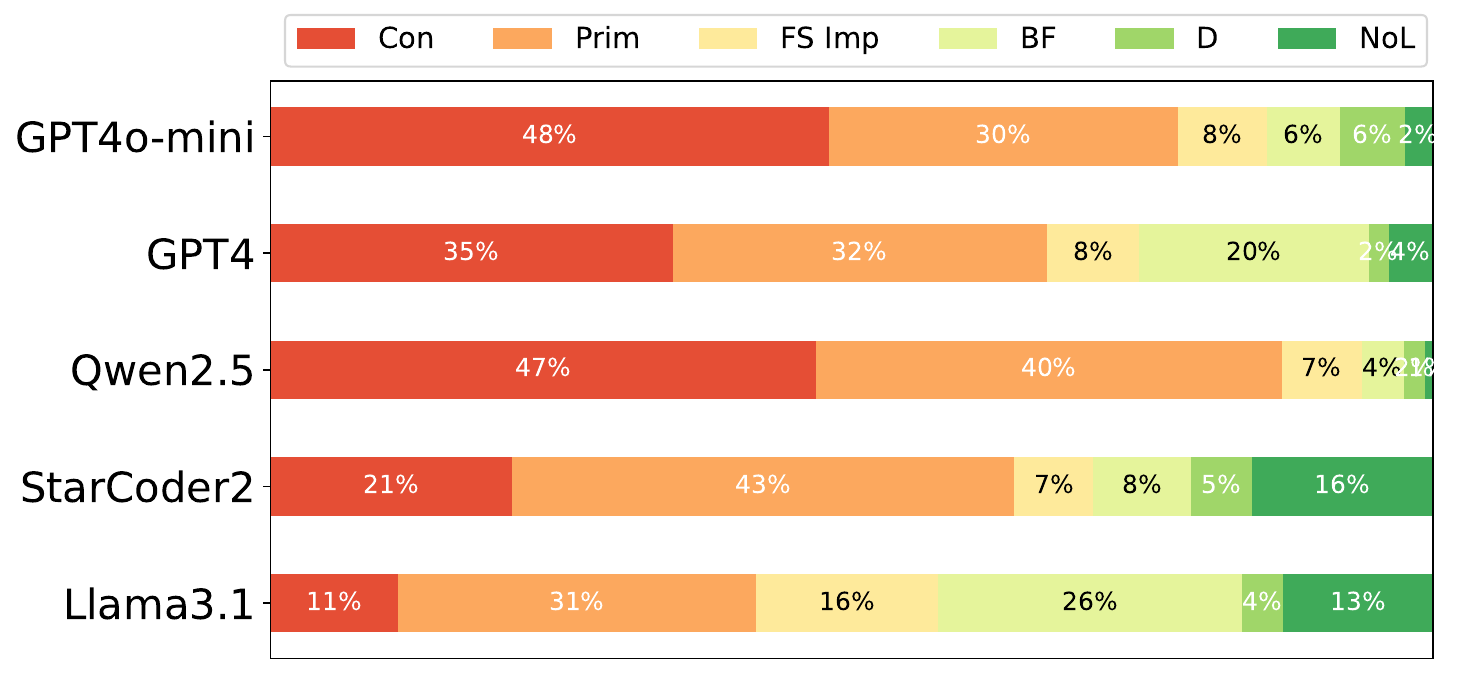}
  \caption{Distribution of programs logic for all evaluated models in percentages on ASDiv problems. Classes are: 'Con': Conceptual, 'Prim': Primitives, 'FS Imp': From scratch Implementation, 'BF': Brute-Force, 'D': Disorganized, and 'NoL': No Logic.}
  \label{fig-classes-distripution}
\end{figure}

We employ Code-Structure judge for automatically evaluating the entire set of generated programs in response to various math problems, and provide the findings below: 

\paragraph{LLMs' capabilities significantly impact the type of implemented reasoning.}
The majority of programs generated by GPT models and Qwen implement logically sound reasoning to solve input problems from the ASDiv dataset. These LLMs ground their programs in mathematical concepts, utilizing numerous API calls to math libraries. Additionally, for many questions, they employ only primitive operations, given the simplicity of some problems in the ASDiv dataset. Figure \ref{fig-classes-distripution} illustrates this, where for the above-mentioned models, the Conceptual and Primitive classes are dominant in the distribution of all programs.
On the other hand, the open-source models, StarCoder2 and Llama3.1, rely much more on Primitve programs but also on ungrounded, unsound reasoning hacks to implement the solution. For instance, Llama3.1 employs math libraries in only 11\% of its programs, while resorting to Brute-Force searches and memorized information, demonstrated in No Logic class, in almost 40\% of its generated programs.

\begin{figure}[t]
  \includegraphics[width=\columnwidth]{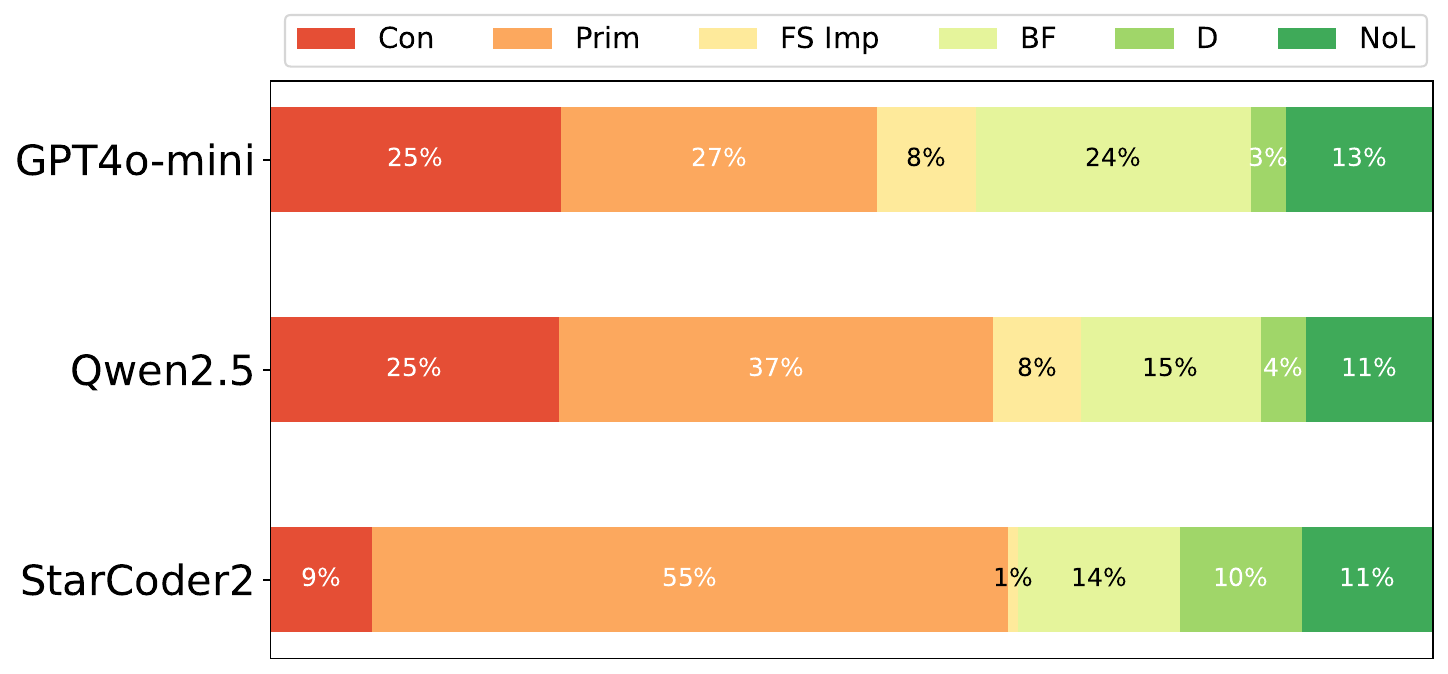}
  \caption{Distribution of programs logic for all evaluated models in percentages on the MATH500 dataset. Classes are: 'Con': Conceptual, 'Prim': Primitives, 'FS Imp': From scratch Implementation, 'BF': Brute-Force, 'D': Disorganized, and 'NoL': No Logic.}
  \label{fig:MATH-classes-distripution}
\end{figure}

\paragraph{Complex math problems increase the generation of unsound reasoning for all evaluated LLMs.}
\label{asdiv-math-comparison}
The difficulty imposed by the MATH500 dataset completely alters the type of reasoning implemented in the generated programs.
This dataset is reported to be challenging for LLMs \cite{math}, as it probs for skills such as Calculus, Geometry, Algebra, Probability, among others.
GPT4o-mini and Qwen now generate 25\% more programs with unsound, ungrounded reasoning, specifically, with exhaustive searches and programs lacking any logic, at the expense of Conceptual programs that utilize math libraries. Figure \ref{fig:MATH-classes-distripution} demonstrates this phenomenon in the distribution of all generated programs. Upon qualitatively analyzing some programs generated by GPT-4o-mini in the No Logic class, we found that the drastic increase in the number of programs with this class can be attributed to the increase in programs with reasoning in the comments rather than code lines. 
We classify these instances as No Logic, because they resemble reasoning in natural language, and the code is not efficiently helping in any way to find the correct answer. The preference of textual reasoning over code might be due to problem complexity, as empirically investigated and discussed in \citep{chen2025steering}, demonstrating that some GPT models prefer textual reasoning over code reasoning depending on the complexity of the task.

\subsection{Further Analyses} 
In this section, we study the impact of sound reasoning, or its absence, on LLMs' end performance. Furthermore, we demonstrate how different math subdomains impact LLMs' preferred logic for solving the problem. 

\paragraph{Impact of type of implemented reasoning on LLMs End-performace.}
To investigate the impact of the reasoning implemented in generated programs on LLMs' end-performance on math tasks, we execute the generated programs and match the execution outcome to ground truth answers, then calculate execution accuracy over programs in each logic class. Table \ref{table:MATH-exec-acc} presents execution accuracy per class on all datasets.
On ASDiv, we observe that sound programs, from Conceptual, Primitive, and From-Scratch Implementation classes, score slightly higher accuracy than unsound programs. However, StarCoder2 and Llama3.1 fail to follow the same trend. Qualitative analysis of their generated programs indicates that while these two LLMs employ API calls or from-scratch implementation to solve the problems, they call or implement the wrong API functionalities, causing the observed low accuracy.  
On the challenging dataset MATH500, Table \ref{table:MATH-exec-acc} illustrates that execution accuracy of programs is on par for all logic types, sound and unsound. This is problematic, as programs with logically unsound reasoning, \textit{i.e:}, false positives, can't be trusted or easily verified; instead, LLMs seem to hack their way to finding final answers.
Finally, the reported high accuracy on the Disorganized programs is mainly due to some false positive predictions from the Code-Structure judge. 

\begin{table*}[t]
  \centering
  \begin{tabular}{lcccccc}
    \toprule
    \multicolumn{7}{c}{\textbf{ASDiv}} \\
    \midrule
    
    \textbf{Model} & \textbf{Conceptual} & \textbf{Primitive} & \textbf{FS Imp} & \textbf{Brute-Force} & \textbf{Disorganized} & \textbf{No Logic} \\
    \midrule
    
    GPT4o-mini   & \(90\%\)  & \(96\%\)  & \(86\%\)  & \(77\%\)  & \(93\%\)  & \(85\%\)
    \\
    GPT4        & \(86\%\)  & \(86\%\)  & \(100\%\) & \(68\%\)  & \(60\%\)  & \(63\%\)  \\
    
     Qwen2.5     & \(91\%\)  & \(78\%\)  & \(89\%\)  & \(100\%\) & \(60\%\)  & \(100\%\) \\
     
    StarCoder2          & \(48\%\)  & \(44\%\)  & \(76\%\)  & \(47\%\)  & \(30\%\)  & \(61\%\)  \\
    Llama3.1     & \(35\%\)  & \(48\%\)  & \(37\%\)  & \(53\%\)  & \(33\%\)  & \(57\%\)  \\ 
    \midrule
    \multicolumn{7}{c}{\textbf{MATH500}} \\
    \midrule 
    
    GPT4o-mini & \(54\%\) & \(57\%\) & \(41\%\) & \(46\%\) & \(64\%\) & \(51\%\) \\
    Qwen2.5   & \(37\%\) & \(59\%\) & \(51\%\) & \(52\%\) & \(23\%\) & \(56\%\) \\
    StarCoder2   & \(0\%\) & \(15\%\) & \(0\%\) & \(25\%\) & \(0\%\) & \(38\%\) \\
    
    \bottomrule   
  \end{tabular}
  \caption{\label{table:MATH-exec-acc}
    Execution (macro) accuracy in percentages per logic class for evaluated models on ASDiv (top) and MATH500 (bottom). Despite the unsound reasoning implemented in programs from Brute-Force and No Logic, high execution accuracy can still be achieved. High accuracy on Disorgnized programs is mainly due to false positive predictions from the Code-Structure judge. 'FS Imp' is From-Scratch Implementation.
  }
\end{table*} 

\begin{figure}
    \begin{subfigure}[b]{0.48\textwidth}
      \includegraphics[width=\linewidth]{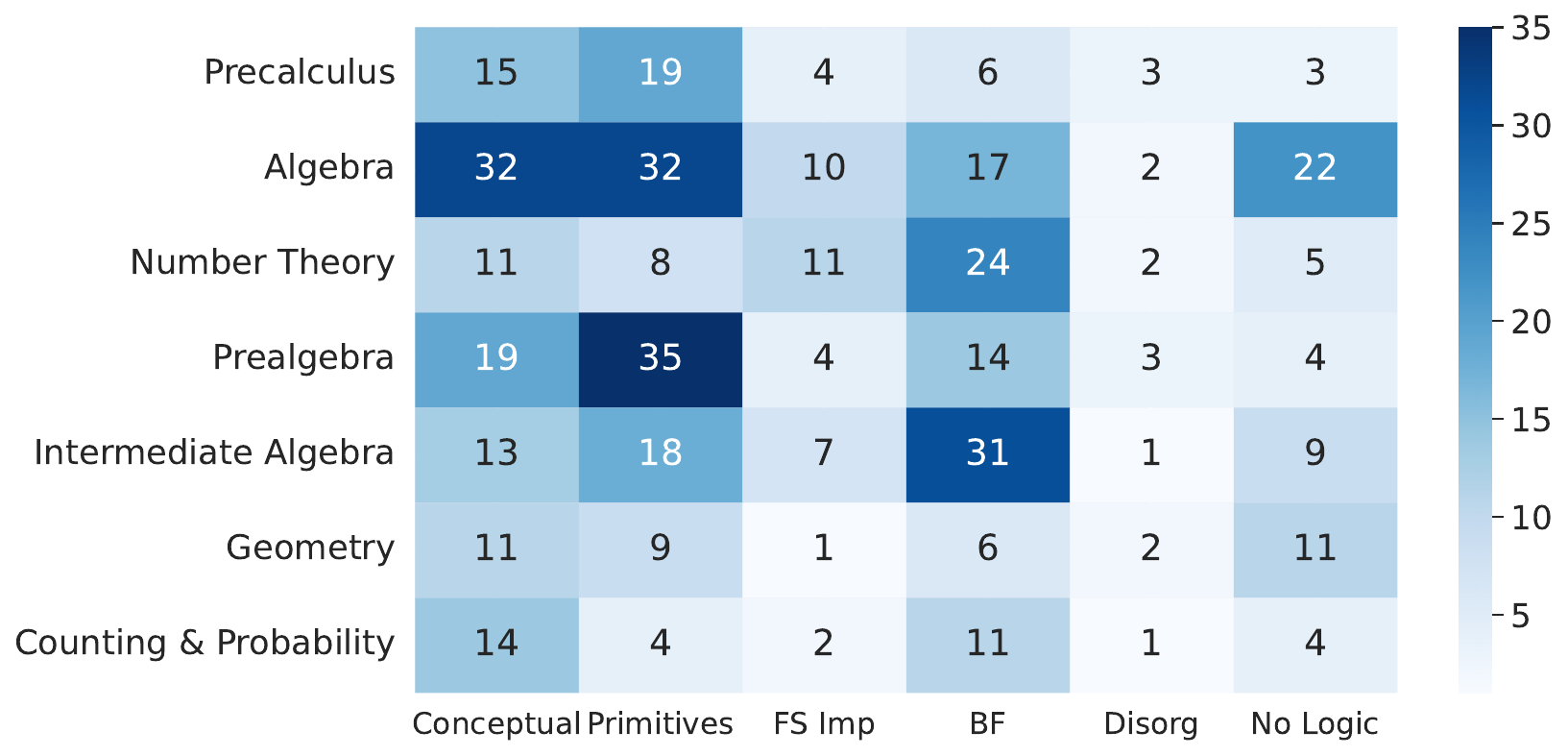}
      \caption{GPT4o-mini}
    \end{subfigure}
    \begin{subfigure}[b]{0.48\textwidth}
      \includegraphics[width=\linewidth]{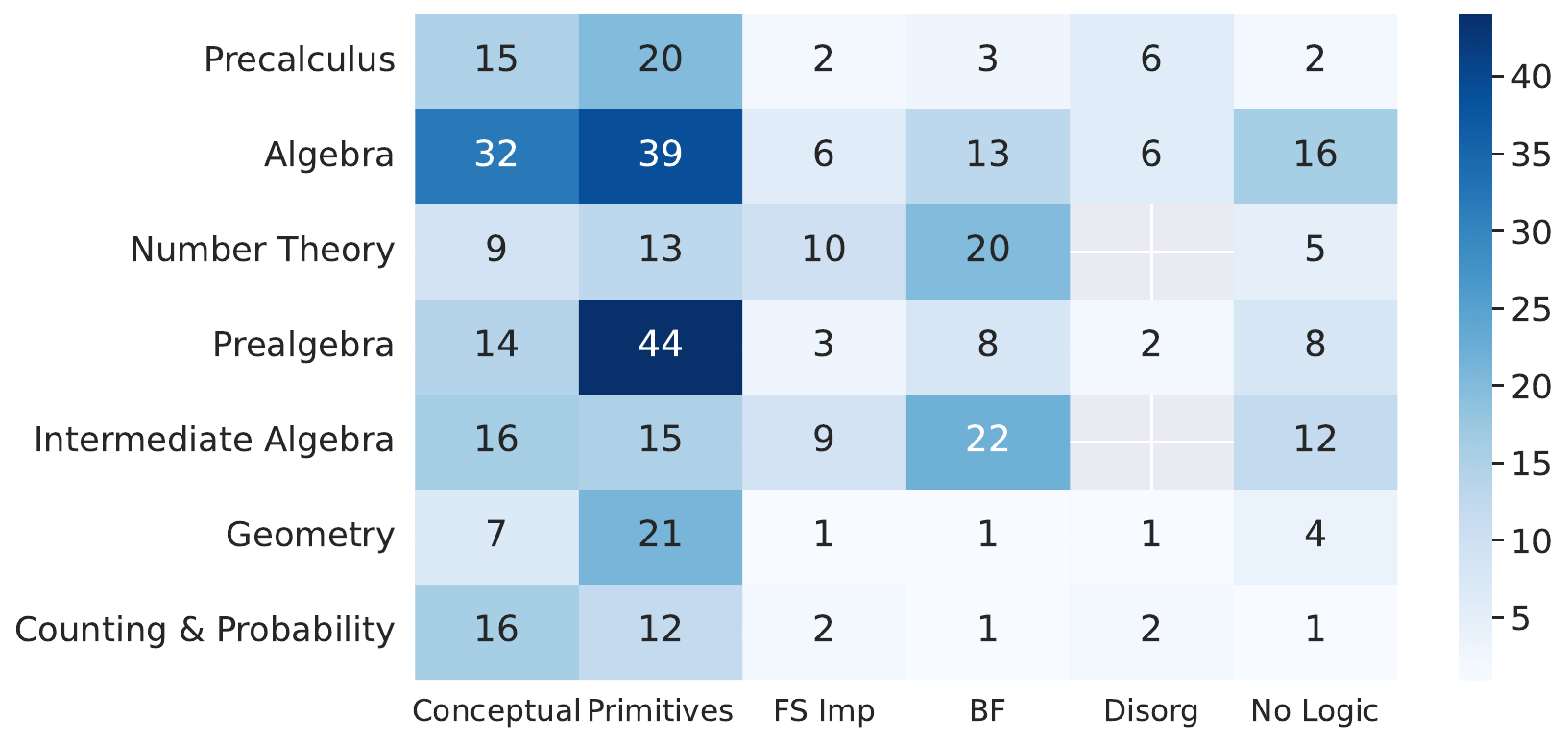} 
      \caption{Qwen2.5}
    \end{subfigure}
    \begin{subfigure}[b]{.48\textwidth}
      \includegraphics[width=\linewidth]{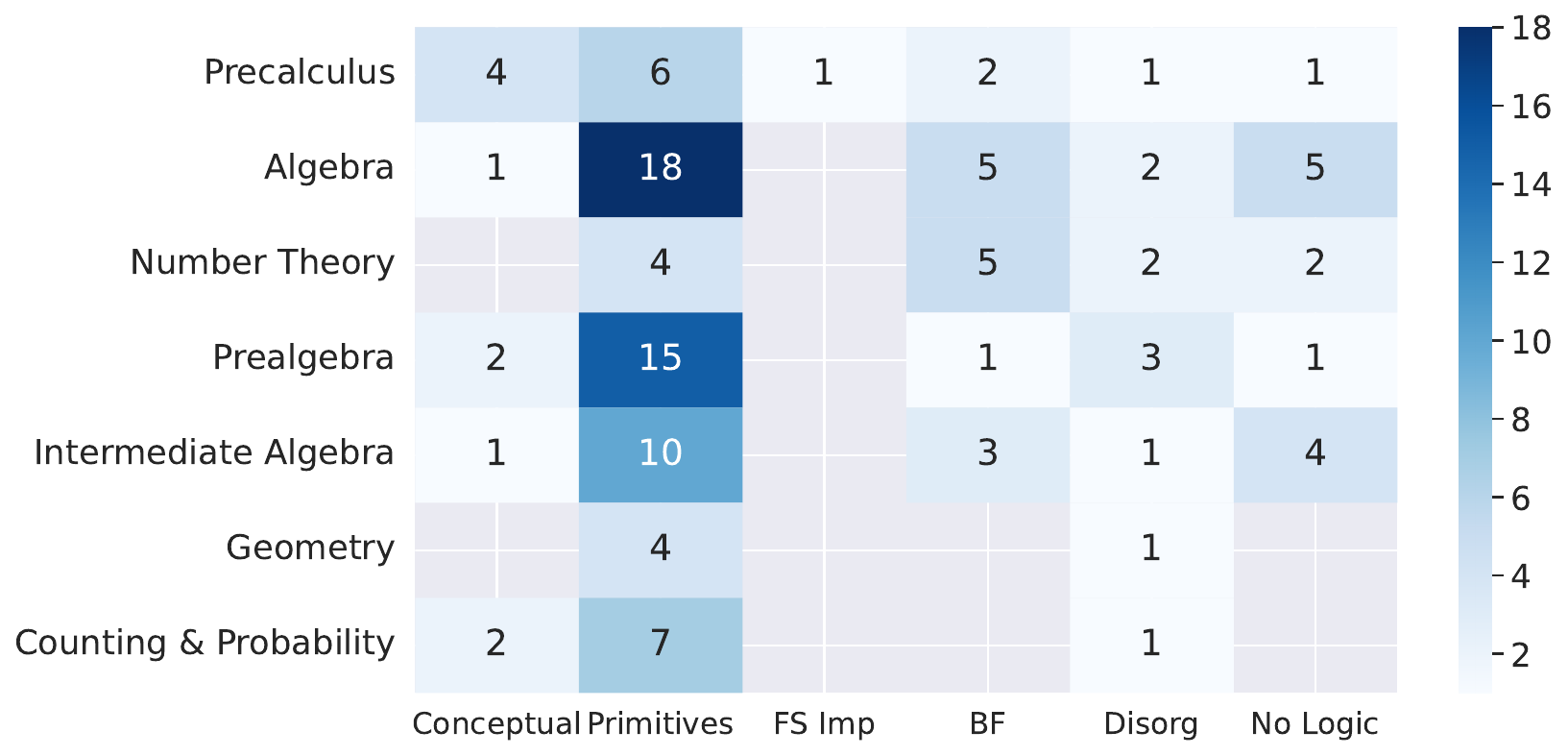}
      \caption{StarCoder2}
    \end{subfigure}
      \caption {Distribution of programs logic, showing counts, per math subdomain in MATH500. Classes: 'FS Imp': From-Scratch Implementation, 'BF': Brute-Force, 'Disorg': Disorganized.}
      \label{fig-skill-class}
\end{figure}

\paragraph{Impact of problem domain on type of implemented reasoning.}
Given that MATH500 questions are annotated with the math subdomain they test for, such as Calculus, Algebra, Probability, etc, we investigate whether the evaluated LLMs consistently approach problems within a domain using the same logic.
We observe that GPT4o-mini and Qwen2.5 tend to use more Primitive solutions for easier problems, such as Prealgebra. Additionally, they consistently employ Conceptual programs for Algebra problems. 
However, both models appear to be less consistent with the type of reasoning they employ across the rest of the subdomains, indicating higher uncertainty about the best logic to tackle the problems.
Figure \ref{fig-skill-class} illustrates the distribution of programs' logic per MATH500 subdomains.
StarCoder2, on the other hand, heavily relies on Primitive solutions for all subdomains, demonstrating a lack of diversity in the logic it employs for different types of math problems. 


\section{Conclusion}
In this work, we conducted an in-depth analysis of code-assisted LLMs generated programs in response to math problems. Our assessment focuses on evaluating the logical soundness of underlying reasoning processes implemented in the generated programs. 
We categorized the generated programs into six different categories, which make up our proposed taxonomy of logical soundness. Three of which represent sound reasoning that ground the programs in verifiable math concepts. In contrast, the other three exploit memorized information or exhaustive searches to find final answers.
We annotated a subset of programs using classes from the taxonomy and trained a decision-tree classifier, which we call the Code-Structure Judge. The Code-Structure judge outperformed an LLM-judge baseline and was employed for a large-scale evaluation of more generated programs. Our findings show that the capabilities of LLMs and the difficulty of the problem impact the type of reasoning implemented, yet regardless, LLMs can exploit unsound reasoning to achieve comparable accuracy. 
Our work underscores the importance of a comprehensive evaluation of code-assisted LLMs. We hope our findings inspire future work to further study why LLMs employ ungrounded solutions and how to mitigate this phenomenon.

\section*{Limitations}
We note there are limitations to our work. 
First: Code-structure Judge is still not accurate on the Disorganized class, causing many false positives.
Second, we depend on one prompting technique and don't compare performance when utilizing prompts to improve generated programs with further refinement, such as Self-Refine \citep{madaan2024self} or Code-based Self-Verification \citep{zhou2023solving}.
   
\section*{Acknowledgments}
The authors would like to thank Vagrant Gautam for their mentorship and useful feedback. We are also grateful to Marius Mosbach and Ellie Pavlick for feedback during the early stages of this work, to Miaoran Zhang for the valuable discussions, and to Tural Mammadov for help with the experimental infrastructure. We also thank anonymous reviewers for feedback. The authors received funding from the DFG (German Research Foundation) under project 232722074, SFB 1102.

\bibliography{paper}

\appendix

\section{Experimental Details}
\subsection{Implementation Details}\label{Impdet}
We use \texttt{starcoder2-15b-instruct}, \texttt{Meta-Llama-3-8B-Instruct}, \texttt{Qwen2.5-7B-Instruct} from HuggingFace \cite{wolf-etal-2020-transformers}. For OpenAI models we use  \texttt{gpt-4-32K} and \texttt{gpt-4o-mini}
We use int8bit model quantization for all models except OpenAI models as we observe no major differences in the execution accuracy per model.
Finally, we use NVIDIA A100 GPUs 40GB, with batch size = 32 for evaluating Llama3 and StarCoder2 and Qwen2.5 on the ASDiv, which took around one hour. While evaluating these models on the MATH subset took about 3 hours with batch size=16. 

\subsection{Resolving some execution errors}\label{res exec errors}
To resolve import errors we used \texttt{autoimport} \citep{autoimport}. For
indentation errors we used the \texttt{inspect} module from the Python
standard library as a post processing step.

\subsection{PAL Prompt and MATH Prompt}\label{demosdet}
In our experiments, we employ PAL prompt from \citep{pal} to evaluate LLMs on the ASDiv data, however we only 
include three demonstrations out of eight, as we observe no further gains from including the entire set. The full prompt is in Figure \ref{fig-pal-prompt}.

\begin{promptlisting}{fig-pal-prompt}{PAL prompt used to prompt evaluated models with three demonstrations on the ASDiv dataset.}
	System: \texttt{Write python program to solve math problems.}
\\
		\, \\
	User: \texttt{\# Q: Michael had 58 golf balls. On tuesday, he lost 23
		golf balls. On wednesday, he lost 2 more. How many golf balls
		did he have at the end of wednesday? } \\
\,
	\\
	Assistant:
\begin{lstlisting}
    # solution in Python:
    def solution():
	"""Michael had 58 golf balls. On tuesday, he lost 23 golf balls.
	On wednesday, he lost 2 more. How many golf balls did he have at
	the end of wednesday?"""
        golf_balls_initial = 58
        golf_balls_lost_tuesday = 23
        golf_balls_lost_wednesday = 2
        golf_balls_left = golf_balls_initial \
		- golf_balls_lost_tuesday \
		- golf_balls_lost_wednesday
        result = golf_balls_left
        return result
\end{lstlisting}

\, \\

	User: \texttt{\# Q: Olivia has \$23. She bought five bagels for \$3
		each. How much money does she have left?} \\
		\,
		\\

	Assistant:
\begin{lstlisting}
    # solution in Python:
    def solution():
	"""Olivia has $23. She bought five bagels for $3 each. How much
	money does she have left?"""
        money_initial = 23
        bagels = 5
        bagel_cost = 3
        money_spent = bagels * bagel_cost
        money_left = money_initial - money_spent
        result = money_left
        return result
\end{lstlisting}

\, \\

	User: \texttt{\# Q: There were nine computers in the server room. Five
		more computers were installed each day, from monday to
		thursday. How many computers are now in the server room?} \\
		\, \\

	Assistant:

\begin{lstlisting}
    # solution in Python:
    def solution():
	"""There were nine computers in the server room. Five more
	computers were installed each day, from monday to thursday. How
	many computers are now in the server room?"""
        computers_initial = 9
        computers_per_day = 5
        num_days = 4  # 4 days between monday and thursday
        computers_added = computers_per_day * num_days
        computers_total = computers_initial + computers_added
        result = computers_total
        return result
\end{lstlisting}
\end{promptlisting}

Th prompt employed for evaluating LLMs on the MATH data is found in Figure\ref{fig-math-prompt}
\begin{promptlisting}{fig-math-prompt}{Prompt structure for evaluating on MATH dataset.}
System: \texttt{Write python programs to solve math problems} \\
User: \texttt{ \# Q: Find the coefficient of \(x^3\) when \(3(x^2 - x^3+x) +3(x +2x^3- 3x^2 + 3x^5+x^3) -5(1+x-4x^3 - x^2)\) is simplifie.
} \\
Assistant:
\begin{lstlisting}
# solution in Python:
from sympy import symbols, simplify
def solution():
    x = symbols('x')
    expr = 3*(x**2 - x**3 + x) + 3*(x + 2*x**3 - 3*x**2 + 3*x**5 + x**3) -
      5*(1 + x - 4*x**3 - x**2)
    simplified_expr = simplify(expr)
    x3_coefficient = simplified_expr.as_coefficients_dict()[x**3]
    result = x3_coefficient
    return result
\end{lstlisting} \\
User: \texttt{\# Q: The surface area of a sphere with radius r is \(4\pi r^2\). Including the
  area of its circular base, what is the total surface area of a hemisphere with
  radius 6 cm? Express your answer in terms of pi} \\
Assistant:
\begin{lstlisting}
# solution in Python:
import math
def solution():
    radius = 6
    # Surface area of the hemisphere
    hemisphere_area = 2 * math.pi * radius**2
    # Area of the circular base
    base_area = math.pi * radius**2
    # Total surface area
    total_surface_area = hemisphere_area + base_area
    result = total_surface_area / math.pi
    return result
\end{lstlisting} \\
User: \texttt{
  \# Q: Monica tosses a fair 6-sided die.  If the roll is a prime number, then
  she wins that amount of dollars (so that, for example, if she rolls 3, then
  she wins 3 dollars).  If the roll is composite, she wins nothing. Otherwise,
  she loses 3 dollars. What is the expected value of her winnings on one die
  toss? Express your answer as a dollar value to the nearest cent.
} \\
Assistant:
\begin{lstlisting}
# solution in Python:
def solution():
    # Probabilities of each outcome
    prime_prob = 1 / 6
    composite_prob = 1 / 3
    otherwise_prob = 1 / 6
    # Expected value of each outcome
    prime_expected_value = (2 * prime_prob) + (3 * prime_prob) + (5 * prime_prob)
    composite_expected_value = 0 * composite_prob
    otherwise_expected_value = -3 * otherwise_prob
    # Total expected value
    total_expected_value = prime_expected_value + composite_expected_value +
      otherwise_expected_value
    # Dollar value to the nearest cent
    result = "{:.2f}".format(total_expected_value)
    return result
\end{lstlisting}
\end{promptlisting}

\subsection{LLM-Judge Prompt}\label{app-llm-judge}
Figure \ref{fig-llm-judge} present the prompt used to prompt o3-mini as an LLM-Judge
\begin{promptlisting}{fig-llm-judge}{The prompt for the LLM-Judge task with o3-mini}
	User: \begin{verbatim}
You are an expert code judge that analyze code based on the
following classes: 
 1- Conceptual through library calls. Reference a math concept
 through calls to relevant math libraries, standard or external. 
 2- primitive solution: programs are expressed in terms of the
 primitive operations only due to problem simplicity, where no
 library functionality can be called or implemented. 
 3- From-scratch Implementation of a library functionality.
 Implements a library functionality from scratch. implementation
 is inlined in the generated code, or can be a custom function
 to be called when required.
 4- Brute-Force. The program search through all possible values
 to find the answer without guiding the search with some math
 knowledge.
 5- Disorganized: the program consists of incoherent steps that
 seem to be a mix of the previous classes. Usually include
 variables used but not defined or the opposite. 
 6- No Logic: These programs merely return a result without
 explicitly generating the steps to arrive at it, transcribing
 information from the question only without further processing
 the information is also No logic. generating the logic as
 comments doesn't count either.
\end{verbatim} \\ 
\begin{verbatim}
Instructions:
- given an input program your task is to analyze it and then
provide a class number from the list above.
- don't fix the code.
- Put your final answer in \boxed{}.
\end{verbatim}
\end{promptlisting}

\section{Training Set Annotation Guidelines and Annotators Background}\label{sec:annotation-guidelines}
Double independent annotation was performed on 50 of the 300 samples, with
\(93\%\) line-granularity and \(100\%\) sample-granularity agreement.
Disagreements were then adjudicated, and guidelines were updated to resolve the
ambiguities.
\subsection{Annotation format}

Each line of program code is prefixed with five characters of the form
\texttt{TPIR} plus space.

Columns:
\begin{itemize}
	\item[\texttt T.] transcription
	\item[\texttt {P1-5}.] processing (blank implies "no logic")
	\item[\texttt I.] inference-time computation
	\item[\texttt R.] result collection
\end{itemize}

\subsection{Transcription}

Purely transcriptive statements that transcribe either data (e.g. numbers)
or relations between data (e.g. equations) from the question.

Purely transcriptive statements can still contain operations that are
explicitly described in the problem statement; these operations do not count as
processing.

\begin{verbatim}
     """ One number is twice as
     large as another. The smaller
     number is 3. Find the other
     number """
     def solution():
T      x = 3
T  R   y = 2*x
   R   return y
\end{verbatim}

Even if the model uses the question to compute at inference time, without
pure transcription of data or relations no \texttt T is marked:

\begin{verbatim}
     """
     One number is twice as large
     as another. The sum of the
     numbers is 12. Find the gcd of
     the two numbers
     """
     def solution():
  I    x = 4
  I    y = 8
 1 R   z = math.gcd(x, y)
   R   return z
\end{verbatim}

If the model has performed partial computations (collapsing e.g. a pure
transcription and true processing) then it receives no \texttt T; it can
possibly receive other labels.

\begin{verbatim}
     """
     Two boards have total length
     10; the long board is 2 longer
     than the short board. Find the
     lengths of the two boards
     """
     def solution():
T      total_length = 10
T      difference = 2 
 2IR   short_length = (total_length - 
         difference) / 2 
T  R   long_length = short_length +
         difference
   R   return short_length, long_length
\end{verbatim}

\subsection{Processing types}

Comments are never marked as anything!!

\begin{itemize}

	\item[1.] conceptual lib
		\begin{itemize}
			\item calls a library function later
			\item mark the entire function; but not the comments
		\end{itemize}

	\item[2.] primitive
		\begin{itemize}
			\item uses primitives in a way that is correct, and cannot be simplified by lib
		\end{itemize}

\item[3.] from scratch implementation
	\begin{itemize}
		\item plausibly correct, looks like inlined implementation of a lib function
		\item writes a function for itself, then calls (mark the call as implementation as well.)
	\end{itemize}

\item[4.] brute-force
	\begin{itemize}
		\item search through all space
		\item mark the entire loop
	\end{itemize}

\item[5.] disorganised
	\begin{itemize}
		\item probably incorrect (more or less random?, disorganised)
	\end{itemize}
\end{itemize}

Empty 'processing type' field (i.e., a space " ") indicates that no processing
occurs.

\subsection{Inference-time computation}

If the model skips steps (either entirely, or in part (in which case there will
still be a processing label)) Then the line gets the "inference-time
computation" flag

In the case of the model entirely precomputing, the processing type will
probably be empty and the \texttt I flag will appear after.

\begin{verbatim}
     def solution():
  I    x = 6 # lcm of 3 and 2
       return x
\end{verbatim}

If the model has performed symbolic manipulation to avoid the use of symbolic
equation libraries, it receives \texttt I as processing/transcription labels.

\subsection{Result collection}

Lines that return the final answer, or that store the variable holding the
final answer, are marked \texttt R in the final column.

\begin{verbatim}
     def solution():
  IR   x = 6 # lcm of 3 and 2
   R   return x
\end{verbatim}

\begin{verbatim}
     """ One number is twice as
     large as another. The sum of
     the numbers is 12.  Find the
     gcd of the two numbers"""
     def solution():
  I    x = 4
  I    y = 8
 1 R   z = math.gcd(x, y)
   R   return z
\end{verbatim}

\subsection{Annotators Background}
The annotation process of the generated programs is time-consuming and isn't feasible to do at a larger scale because it requires expertise with code understanding. Both annotators are experts in computer science and have done prior reading on programming languages and related topics, that make them a better fit for the annotation process.

Furthermore, programmatic steps and code structure patterns are less subjective than natural language, carrying less nuance, which can lead to disagreement or bias. For example, it is hard to mistake a library call from a brute-force program.

\section{More Results on Code-Structure Judge and LLM-Judge Performance}
\subsection{Per-class \(F_1\) scores}
Table \ref{tbl:p-r-f1} provides precision, recall, and F1 scores of both code judges.

\begin{table*}[h]
\centering
\begin{tabular}{l||r|r|r}
    \toprule
         
       & \textbf{Precision} & \textbf{Recall} & \(\mathbf{F_1}\) \\
    \midrule
    Conceptual     &  \(1.\)  &  \(1.\)  &  \(1.\) \\
    Primitive  &  \(0.72\)  &  \(0.95\)  &  \(0.82\) \\
    From-scratch Implementation &  \(0.66\)  &  \(1.\)  &  \(0.8\) \\
    Brute-Force   &  \(0.88\)  &  \(1.\)  &  \(0.94\) \\
    Disorganized &  \(0.5\)   &  \(0.16\)   &  \(0.25\)  \\
    No Logic    &  \(0.91\)   &  \(0.68\)   &  \(0.78\)  \\

\end{tabular}
\begin{tabular}{l||r|r|r}
    \toprule
       & \textbf{Precision} & \textbf{Recall} & \(\mathbf{F_1}\) \\
    \midrule
    Conceptual     &  \(1.\)  &  \(0.88\)  &  \(0.94\) \\
    Primitive  &  \(0.56\)  &  \(0.95\)  &  \(0.71\) \\
    From-scratch Implementation      &  \(0.83\)  &  \(0.83\)  &  \(0.83\) \\
    Brute-Force         &  \(0.83\)  &  \(0.93\)  &  \(0.88\) \\
    Disorganized    &  \(0.6\)   &  \(0.25\)   &  \(0.35\)  \\
    No Logic    &  \(0.75\)   &  \(0.37\)   &  \(0.5\)  \\
\end{tabular}
    \caption{Precision, recall, and \(F_1\) of the Code-Structure Judge (top) and LLM-Judge (bottom)}
    \label{tbl:p-r-f1}
\end{table*}

\subsection{Examples of incorrectly classified programs}\label{dc-wrong-classified} 
We provide a few examples of programs where Code-Structure judge misclassifies Disorganized programs and their true label in Figure \ref{fig-false-positive}.
\begin{figure}
    \centering
    \begin{tabular}{p{0.8\textwidth}}
    \begin{lstlisting}
    def find_other_number(): 
        difference = 100 
        one_number = 91 
        other_number = one_number + difference 
        return other_number 
    result = find_other_number()
    \end{lstlisting}

	\qquad   {\tiny \textbf{True label:} ``Primitive''} \\
    \begin{lstlisting}
    def solution():
        cards_per_pack = 20
        envelopes_per_pack = 17
        # least common multiple of 20 and 17 is 340
        lcm = 340
        result = lcm
        return result
    solution()
    \end{lstlisting}
	    \qquad {\tiny \textbf{True label:} ``No Logic''} \\
    \end{tabular}
    \caption{Instances that were labeled ``Disorganized'' by the Code-Structure Judge, and their true label.}
    \label{fig-false-positive}
\end{figure}

\section{Further Analysis of Generated Programs}
\subsection{Most common bugs in the generated programs on MATH500 dataset}\label{app-code-errors}

StarCoder2 generated 60 programs with bugs: 23 of them with undefined symbols and  15 that were calling undefined functionality from libraries. e.g, calling 'log3' from the math library. 
Qwen2.5 generated 40 programs with bugs as follows: 8 were with undefined symbols, 7 with undefined attributes of objects e.g, a float object has no attribute denominator. 
GPT4o-mini generated only 5 programs with bugs, two of which have undefined symbols.

\end{document}